\documentclass[conference]{IEEEtran}
\IEEEoverridecommandlockouts
\usepackage{times}
\usepackage{epsfig}
\usepackage{graphicx}
\usepackage{subfigure}
\usepackage{multirow}
\usepackage{multicol}
\usepackage{cite}
\usepackage{amsmath,amssymb,amsfonts}

\usepackage{textcomp}
\usepackage{xcolor}

\usepackage{algpseudocode}  

\begin{document}

\title{Interflow: Aggregating Multi-layer Feature Mappings with Attention Mechanism
}

\author{\IEEEauthorblockN{Zhicheng Cai 
}
\IEEEauthorblockA{\textit{School of Electronic Science and Engineering}\\
\textit{Nanjing University},
Nanjing, China 210023\\ Email: 181180002@smail.nju.edu.cn
}}

\maketitle

\begin{abstract}
   Traditionally, CNN models possess hierarchical structures and utilize the feature mapping of the last layer to obtain the prediction output. However, it can be difficulty to settle the optimal network depth and make the middle layers learn distinguished features. This paper proposes the \textbf{Interflow} algorithm specially for traditional CNN models. Interflow divides CNNs into several stages according to the depth and makes predictions by the feature mappings in each stage. Subsequently, we input these prediction branches into a well-designed attention module, which learns the weights of these prediction branches, aggregates them and obtains the final output. Interflow weights and fuses the features learned in both shallower and deeper layers, making the feature information at each stage processed reasonably and effectively, enabling the middle layers to learn more distinguished features, and enhancing the model representation ability. In addition, Interflow can alleviate gradient vanishing problem, lower the difficulty of network depth selection, and lighten possible over-fitting problem by introducing attention mechanism. Besides, it can avoid network degradation as a byproduct. Compared with the original model, the CNN model with Interflow achieves higher test accuracy on multiple benchmark datasets.
\end{abstract}

\section{Introduction}
Since AlexNet \cite{krizhevsky2012imagenet} won the title of ILSVRC-2012, convolutional neural network (CNN) has grown in popularity and influence, increasingly attracted the research interests of the extensive scientific researchers, and flourished in an compelling way. Many novel and high-performance deep CNN models are proposed, which break the test records of various benchmark datasets continuously. The CNN is also applied from the image classification and recognition task like the original handwritten digit recognition \cite{lecun1998gradient} to wide-field, such as object detection \cite{ren2015faster,liu2016ssd,redmon2016you}, semantic segmentation \cite{ronneberger2015u,long2015fully,chen2017rethinking}, image generation \cite{goodfellow2014generative,radford2015unsupervised,zhang2017stackgan}, human posture estimation \cite{alp2018densepose,cao2017realtime,fang2017rmpe} and so on \cite{qi2017pointnet,wang2017deep,simonyan2014two}, all obtaining the achievements. 

On accounting that CNN is a black-box model essentially, ``trial and error'' has become the only approach of choosing the hyper-parameters such as network depth. It is known to us that insufficient convolutional layers leads to a poor model representation ability. However, when the network depth increases to a certain degree, it will lead to issue of network degradation, which means the decline of model representation ability. Although the residual connection alleviates this problem, with the increase of model depth, the extent of the model capacity's improvement is significantly decreased, that is to say, there exists marginal utility phenomenon~\cite{zhang2017polynet}. In addition, the problem of over-fitting can emerge when the number of learned parameters is excessive. As a result, how to choose the number of convolutional layers can be a tricky problem to researchers when fine-tuning and practical application. Besides, it is an outstanding problem that how to make the middle convolutional layers learn more distinguished features, enhancing model representation ability as a result.  

\begin{figure}[htbp]
\centering
\includegraphics[height=0.43\textwidth]{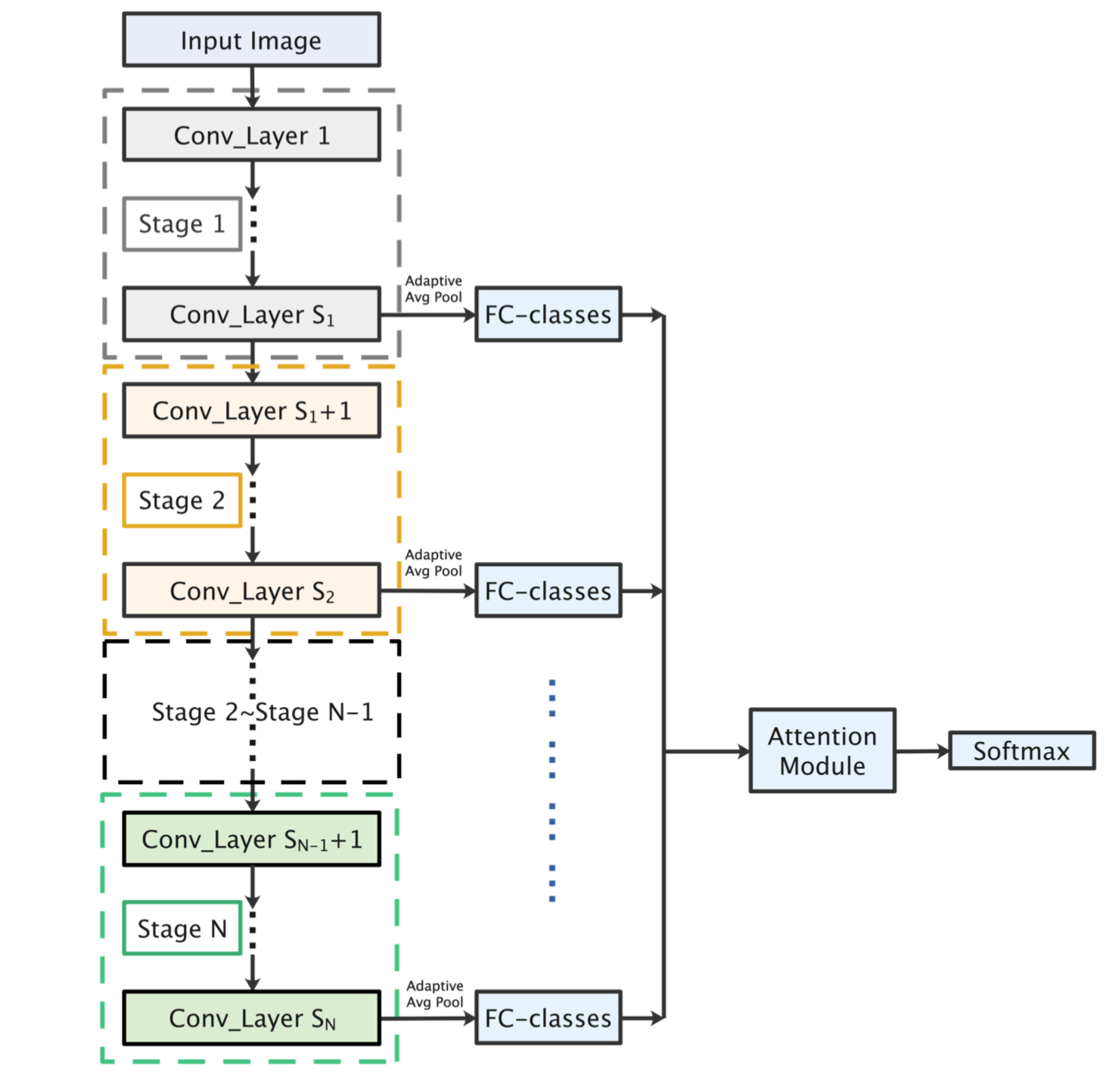}
\caption{Schematic diagram of Interflow algorithm.}
\label{f1}
\end{figure}

Consequently, this paper introduces the \textbf{Interflow} algorithm. It is known to us that CNN model is stacked with convolutional layers, and utilizes the feature mapping of the last layer to obtain the final output. However, Interflow divides the feature mappings of CNN model to various stages and adds a prediction branch following the last convolutional layer of each stage. These branches are aggregated in the well-designed attention module which introduces the attention mechanism to distribute the importance of each branch. Finally, prediction result is obtained as the output of the attention module. This fusion process of the feature mappings deriving from different convolutional stages is similar to the convergence of water flows, thus the algorithm is termed as ``Interflow’’. Fig.~\ref{f1} exhibits the schematic diagram of Interflow implemented in a CNN model.

 Interflow alleviates the difficult of fine-tuning the network depth and forces the convolutional layers in the middle stages to learn more distinguished features. Experimental results validate that Interflow can enhance the representation ability of CNN models. In addition, it utilizes the feature mappings of different convolutional stages efficiently by introducing attention mechanism, which considers the feature mappings of various abstraction levels simultaneously and balances the receptive filed and the resolution. What’s more, Interflow can solve the problem of gradient diminishing for connecting the middle convolutional layers to the final output straightforwardly. As a consequence, the gradient flow can propagate to these learned parameters in lower layers directly through these added prediction branches. 

Moreover, Interflow can deal with the over-fitting problem and alleviate the network degradation phenomenon for that the attention mechanism module is supposed to figure out these valuable or redundant convolutional layers. As a result, it pays more attention to these stages with critical information and allocates light or even zero weight to these unnecessary stages. Later experiments that utilize Interflow on extremely deep CNN models verify this argument as well.

\section{Related Work}
\subsection{Auxiliary Classifier}
Most backbone CNN models only take advantage of the feature mapping in the last layer to perform Softmax classification and output prediction results. However, GoogLeNet~\cite{szegedy2015going} adds two auxiliary classifiers to the middle feature layers (Inception 4a and 4d modules) to aid training. In the training process, the total loss can be obtained by multiplying the classification loss of these two auxiliary classifiers with a light weight of 0.3 and adding these losses to the classification loss obtained by the last layer feature mapping (Inception 5b module). Subsequently, the back-propagation is carried out to update the weights. During the testing process, GoogLeNet simply removes the two auxiliary classifiers and only employs the last classifier for classification and prediction.

The fundamental purpose of adding two auxiliary classifiers to GoogLeNet is to provide regularization and avoid gradient vanishing problem simultaneously. Moreover, it is supposed to conduct the gradient in the forward direction~\cite{szegedy2015going}. In addition, some CNN models with shallow layers achieve relatively good performance, indicating that it is necessary for the middle layers of CNN models to have a strong recognition ability~\cite{szegedy2015going,krizhevsky2012imagenet}. Consequently, the auxiliary classifiers utilize the output of the middle layers to assist classification, forcing the middle convolutional layers to learn discriminative and effective features.

However, the experimental data illustrates~\cite{szegedy2015going} that the auxiliary classifiers make little difference. Compared to GoogLeNet model without auxiliary classifier, GoogLeNet model with auxiliary classifiers only improves the test accuracy by about 0.5\% on ImageNet dataset. Moreover, GoogLeNet model with one auxiliary classifier can achieve almost the same effect. This paper believes that although these auxiliary classifiers GoogLeNet adopted can alleviate the gradient vanishing problem, it fails to utilize the effective features extracted by the middle convolutional layers properly. In addition, the method it fuses the auxiliary classifiers may be improper, thus failing to enable the middle layers to learn more distinguished features. Finally, it directly discards auxiliary classifiers during prediction, as a result, auxiliary classifiers don’t play a positive and effective role in the prediction of the result. 

\subsection{Multi-layer Features}
Consider object detection tasks, the input images often contain both large and small objects. The CNN feature mappings of deep layers possess strong semantic features, which is beneficial to classification and recognition. However, the receptive field is large, while the feature resolution is low~\cite{kong2016hypernet}. In this case, if we still only use feature mappings in the last layer to predict, it can easily give rise to the missed detection of small objects and the low accuracy for positioning bounding boxes. The feature mappings of middle and shallower layers possess larger feature resolution, which can be conducive to locate objects. Therefore, the SSD algorithm~\cite{liu2016ssd} not only sets proposal boxes for the last feature map, but also establishes proposal boxes for the feature mappings of six different scales in the shallow, medium and deep layers. Furthermore, it builds relatively small proposal boxes in the feature map of shallower layers to detect small objects and proposes larger boxes in the feature map of deeper layers to detect large objects.

However, the SSD algorithm has a poor detection effect on small objects. The reason is that the semantic information of the shallow features is insufficient, as a result, it is incapable of completing prediction. Consequently, researchers propose the DSSD algorithm~\cite{fu2017dssd} to perform feature fusion on feature mappings of deep and shallow layers. Specifically, DSSD directly utilizes the feature map of the deepest layer for regression and classification. Sequentially, this feature mapping is deconvoluted and be multiplied by elements with the shallow feature mapping to obtain feature fusion of deep and shallow layers. The output of fusing features can be classified and calculated by regression. Similarly, if we continue to perform deconvolution and fusion of the feature map with the features in shallow layers, we can output a total of 6 feature maps after fusion. These 6 feature maps can be predicted with regression and classification. Therefore, DSSD outputs six predictions.

DSSD integrates the features of deeper layers into the feature mappings of shallower layers, which upgrades the semantic information of the shallow features and improves model performance, especially the detection of small objects. This paper believes that information fusion of deep and shallow features plays a critical role in the performance improvement. However, it is precisely because the feature mappings of different levels employ independent branches of classification and regression prediction that the detection tasks of large objects and small objects can be distinguished. As a result, the detection is preferably completed and these shallower convolutional layers are enabled to learn more discriminant features. Inspired by this, this paper proposes Interflow which adds prediction branches in different layers of backbone CNN models. In the end, through a simple linear weighting or a learned weight layer that intends to achieve the attention mechanism, the predictions of branches are fused to obtain the final output. This design aims at making the network use the abstract information of different levels in the deep and shallow layers simultaneously. In addition, it enables the middle convolutional layers to learn more distinguished features and finally achieves more superior output.

\subsection{Attention Mechanism}
Attention mechanism derives from the study of human vision. In order to efficiently use limited visual resources to process information, humans selectively concentrate on specific parts of the visual area. Attention mechanism is mainly to determine the input parts that need attention and allocate limited information processing resources to these important part~\cite{galassi2019attention}. In the computer vision realm, attention mechanism is introduced to process visual information. Traditional feature extraction of local image~\cite{sharifrazi2021fusion}, saliency detection~\cite{hou2007saliency} and sliding window~\cite{glumov1995detection} can all be regarded as a sort of attention mechanism. Furthermore, the attention mechanism is usually implemented by an additional attention module in neural networks. This attention module can allocate specific weights for different parts of the input. 
Generally speaking, the attention mechanism can be divided into two forms, namely, hard attention mechanism and soft attention mechanism. The hard Attention mechanism employs a manually fixed ``weight mask’’.  It forces the computer to focus on the part requiring attention and reduces the influence from other parts. However, it is only valid for specific sampled pictures. Moreover, since the ``weight mask’’ contains no learned parameters, it is a non-differentiable constant and impossible to update the weight in model. On the contrary, soft attention mechanism employ a layer of learned ``weight mask’’ between the convolutional layers. Consequently, it can recognize the parts that neural networks should pay close attention to through training. The attention mechanism was initially widely used in the field of NLP~\cite{vaswani2017attention,devlin2018bert}. The transformer model which has introduced the self-attention mechanism is a classic case. Inspired by NLP, models in the field of CV have also continuously introduced the attention mechanism~\cite{han2020survey,niu2021review}. 

In CNNs, the Attention Mechanism usually has two implementation patterns, namely, spatial attention and channel attention. They are intended to conduct re-sampling for the input data and strengthen specific objects. In a common model of the spatial attention, the output can be obtained by point-wise matrix multiplying between the differentiable weight mask and the conolutional feature mapping~\cite{wang2018parameter}. On the other side, channel attention carries out weighted calculation for the feature channels. SENet~\cite{hu2018squeeze} is a typical implementation of the channel attention. SENet utilized the SE module to automatically learn the importance degree of each feature channel. CBAM~\cite{woo2018cbam} adopted both spatial and channel attention. Inspired by attention mechanism, Interflow weights the prediction results of feature mappings in different layers by the attention mechanism to get the final output. As a result, the model can learn the importance degree of results obtained by different branches, lower the difficulty of selecting network depth and mitigate network degradation caused by extremely deep networks. 

\section{Interflow}

\subsection{Interflow Algorithm}
According to the above analysis, we are aware that it is necessary for the middle convolutional layers to learn discriminative features. In addition, we are supposed to lower the difficult of choosing the network depth and utilize features in different levels reasonably and effectively. Consequently, the Interflow algorithm is designed for these purposes.

According to the depth of the hierarchy, Interflow first divides the CNN into different stages. For the last convolutional layer at each stage, the size of the feature mapping is reduced to $1\times1$ through an adaptive average pooling. Then, output obtained goes through one fully connected layer to get the confidence coefficients in different categories of this stage. Subsequently, we concatenate these channels possessing varied confidence coefficients of each stage. Equivalently, the feature information from different branches intersects and inputs to the module of the attention mechanism, which aims to allow the model to rationally organize and utilize the features learned at different stages. Therefore, the mode concentrates on valid features and discards redundant features. After aggregation of multi-layer features with attention mechanism, it is finally input to a softmax classifier in order to achieve the final output.
\begin{figure*}[htbp]
\centering
\subfigure[Hard attention module]{
\includegraphics[width=0.47\textwidth]{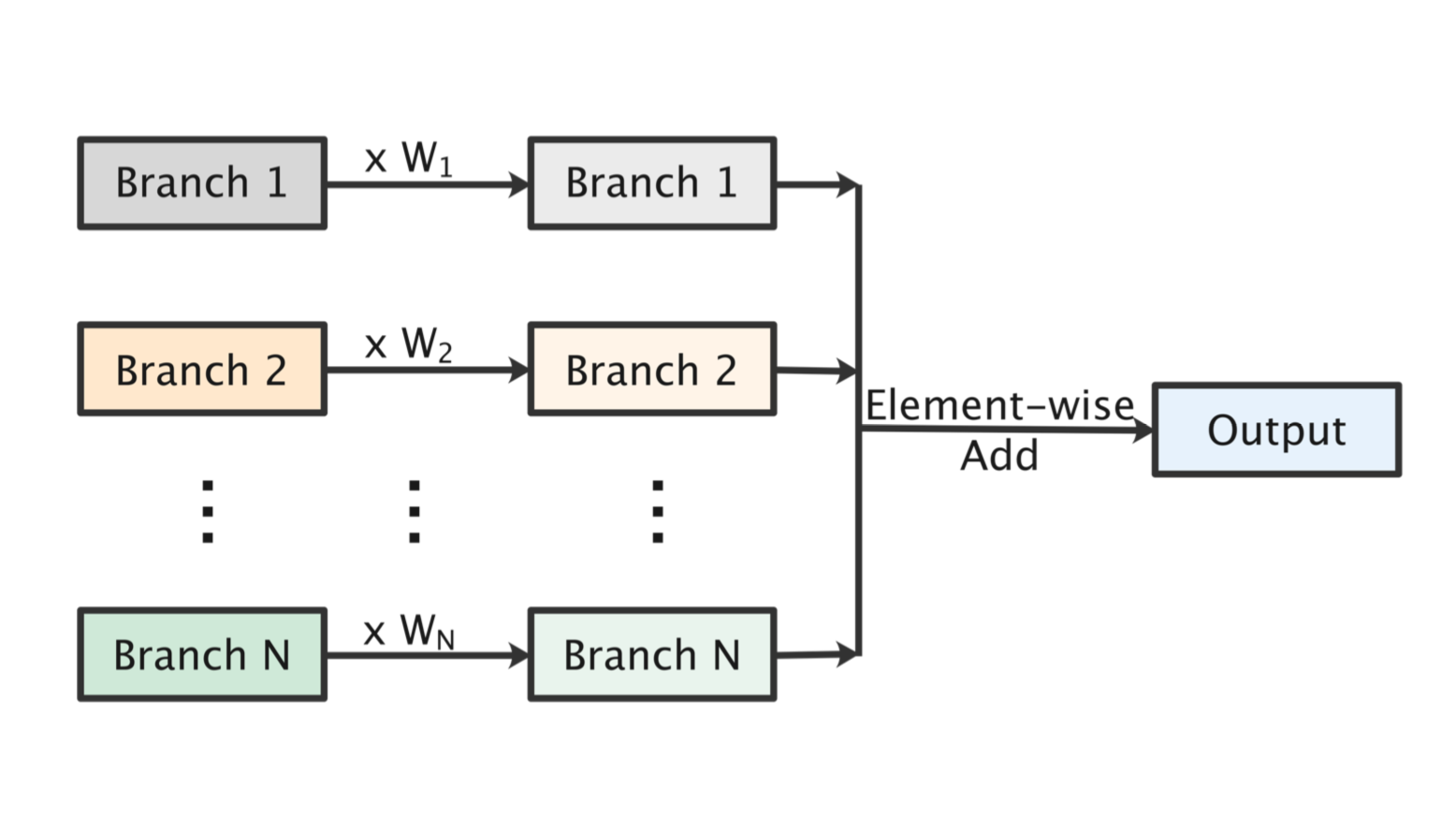}
}
\subfigure[Soft attention module]{
\includegraphics[width=0.47\textwidth]{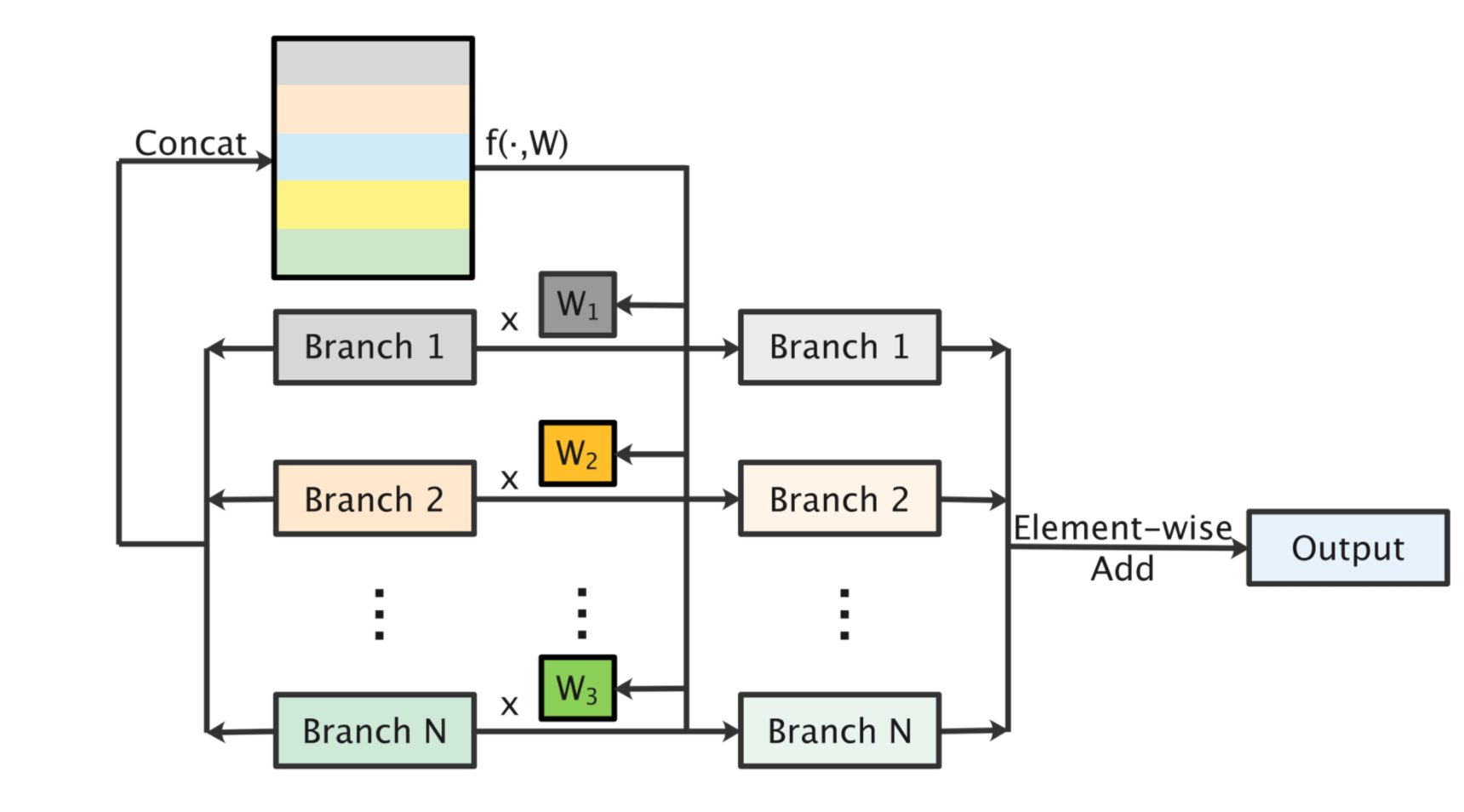}
}
\caption{Schematic diagram of the hard attention module (left) and soft attention module (right).}
\label{f2}
\end{figure*}

\subsection{Attention Module}
As previously mentioned, the implementation methods of attention mechanism include hard attention and soft attention. Similarly, there are two ways to implement modules of Interflow's attention module with the intersection of information flows in various branches. 

\textbf{Hard Attention Module: }As a manually-designed and non-learned weight mask, hard attention is to endow the confidence output of different stages with manually-designed weights. We can straightforwardly regard the weight of each branch as a hyper-parameter. Fig.~\ref{f2} left shows the schematic of hard attention module. However, when the number of branch streams is too large, there will be too many hyper-parameters. Consequently, it is difficult to obtain an optimal combination. In addition, we cannot know which stage the model should focus on for a specific task prior to training it for multiple times. In other words, manual setting is incapable of reasonably integrating the feature information of multiple branches. Therefore, we pay more attention to the Soft Attention.

\textbf{Soft Attention Module :} Soft attention allows the model to learn the weights independently. Specifically, we utilize a $1\times n$ convolution to make the model learn the weight of each branch. Fig.~\ref{f2} right shows the schematic of soft attention module. Therefore, it enables the model to recognize the stage feature information that should be paid attention to for a specific task. In addition, it integrates the feature information of different stages reasonably and effectively. Since the feature information at different stages may have a preference for different categories, namely, it may learn more discriminative features for a specific category. As a result, in subsequent experiments we also make the model to learn the weight of different categories confidence in each extra branch. This is realized by replacing the $1\times n$ convolution with a point-wise $n\times n$ convolution.

\subsection{Advantages}
For clear expression, we summarize the advantages of Interflow below.

\begin{itemize}
    \item By taking advantage of the attention mechanism, Interflow rationally integrates and effectively utilizes the feature mappings from different stages and abstraction levels. Thus enhances the utilization of feature information. Besides, it forces the intermediate convolutional layers to learn more distinguished features.
    \item Interflow can solve the gradient vanishing problem, which is similar to the addition of auxiliary classifiers in GoogLeNet. Certain intermediate convolutional layers are attached to the final output straightforwardly, thus the gradient can be propagated to shallower layers before diminishing.
    \item Moreover, we know that the shallow convolutional layer of CNNs fits faster than the deep convolutional layer[]. Through autonomous learning of branches’ weights at different stages, the branches of high-level phase may have negative weights, while the branches of high-level phase have positive phase weights in the final stage of training, which may slow down the fitting speed in the shallow convolutional layers and cut down the possibility of over-fitting due to over-learning in shallow convolutional layers to a certain extent.
    \item When we design a CNN model, we usually cannot know the reasonable depth of the model in advance. We hope that the Interflow algorithm can learn tiny or even zero weights for the redundant deep-level convolutional layers, thereby decreasing the attention on the feature information of these convolutional layers. As a result, the difficult of choosing network depth is lowered. In addition, it lowers the over-fitting risk of the model due to too many layers or too high complexity.
    \item By allocating slight or zero weights to these extremely deep-level stages which destroy model representation ability, the Interflow algorithm is supposed to alleviate the network degradation problem when applied to extremely deep CNN models.
    \item Interflow enjoys the advantage of portability. It can be utilized in any backbone CNN models.  
\end{itemize}

\subsection{Implementation}
\begin{figure*}[htbp]
\centering
\includegraphics[height=0.76\textwidth]{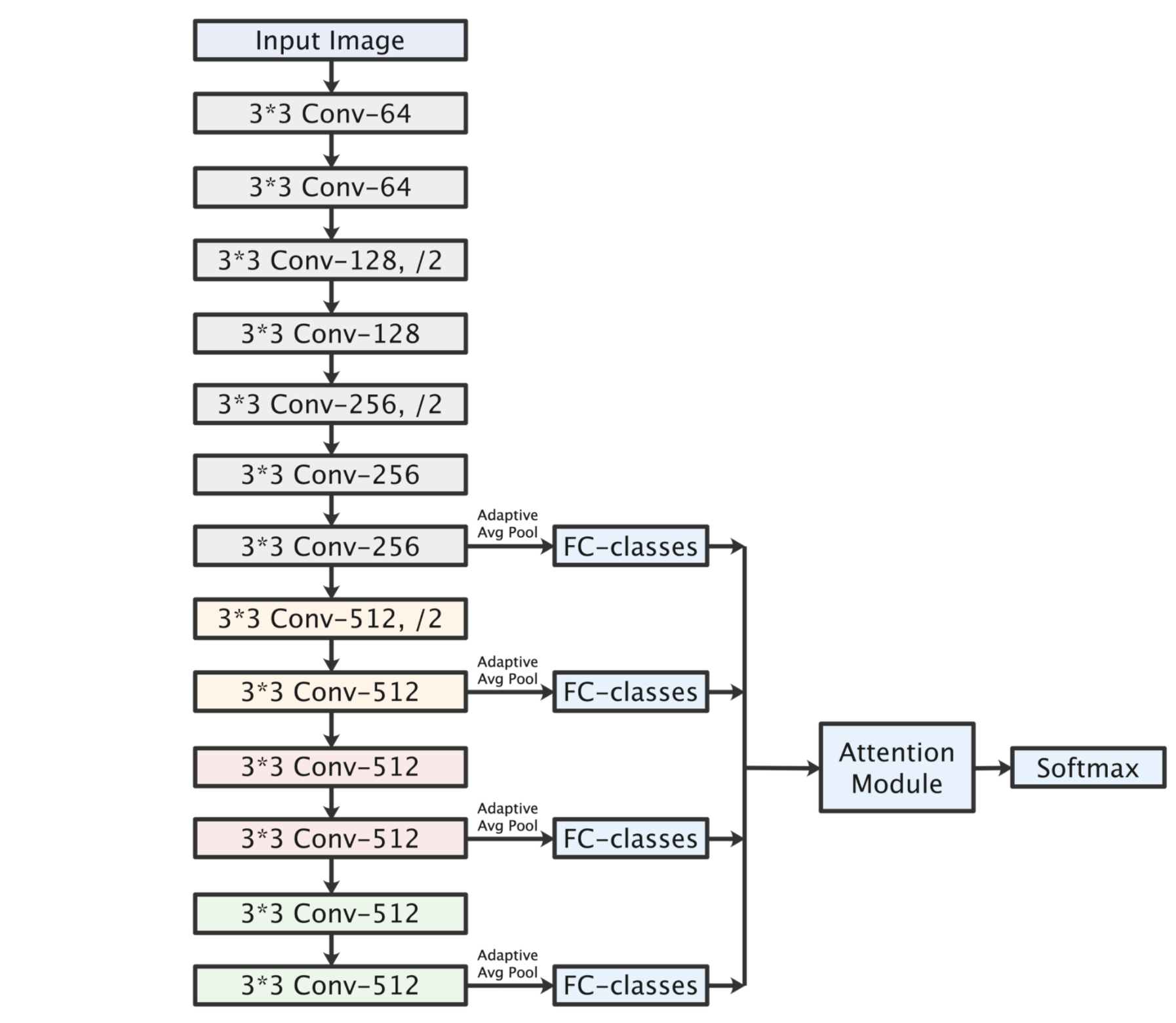}
\caption{Schematic diagram of the CNN model with Interflow.}
\label{f3}
\end{figure*}

Fig.~\ref{f3} is a sketch map of the specific CNN model with Interflow utilized in the subsequent experiments of this paper. In this model, we set the convolutional layer of the VGGNet-16~\cite{simonyan2014very} as the blueprint to extract features. As the figure shows, we divide 13 convolutional layers into 4 stages and proceed each stage’s output feature mappings with an adaptive average pooling and fully connected layers. In this way, we obtain the classification confidence coefficient, namely, the feature information of branches at each stage. We further input the feature map obtained by channel concatenation to the attention mechanism for weighting. Then we gain the final output. In a sense, Interflow can be regarded as a slight model ensemble.

In the application of Interflow, the stage division of the CNN model can be regarded as a hyper-parameter. The basic model utilized in this paper discards convolutional layers from deep to shallow. The results regarding the test accuracy of each model on the CIFAR-10 benchmark dataset are shown in Table.~\ref{t1}, in which ``depth’’ stands for the number of remaining convolutional layers. We can observe that the model has a tolerable expressing ability from the seventh layer. The feature information it has learned can have a certain positive influence on the final prediction, which is also the reason why we divide the first phase into the seventh convolutional layer in the subsequent experiments. Meanwhile, the number of feature channels are same in the 5-th, 6-th and 7-th convolutional layers, while the number of feature channels becomes a constant of 512 since the 8-th layer. There exists an instinctive and obvious fault of the convolution kernel numbers, which is one of the reasons why we set the first stage like this.

\begin{table}[!t]
\renewcommand{\arraystretch}{1.3}
\caption{Test accuracy of CNN models with different depth .}
\label{t1}
\centering
\begin{tabular}{|c|c|}
\hline
Depth & Test Accuracy (\%)\\
\hline
4 Conv-layers& 83.21\\
\hline
5 Conv-layers& 85.85 \\
\hline
6 Conv-layers& 86.41\\
\hline
7 Conv-layers& 87.02\\
\hline
8 Conv-layers& 90.39\\
\hline
9 Conv-layers& 90.53\\
\hline
10 Conv-layers& 90.70\\
\hline
11 Conv-layers& 91.01\\
\hline
12 Conv-layers& 90.40\\
\hline
13 Conv-layers& 91.12\\
\hline
\end{tabular}
\end{table}

\section{Experiment}

\subsection{Dataset}
The benchmark datasets utilized including CIFAR-10, CIFAR-100, SVHN, MNIST, KMNIST, and Fashion MNIST.
\begin{itemize}
\item CIFAR-10: CIFAR-10 dataset is composed of 10 classes of natural images with 50,000 training images in total, and 10,000 testing images. The 10 classes include: airplane, automobile, bird, cat, deer, dog, frog, horse, ship and truck. Each image is a RGB image of size $32\times32$. 
\item CIFAR-100: CIFAR-100 dataset is composed of 100 different classifications, and each classification includes 600 different color images, of which 500 are training images and 100 are test images. As a matter of fact, these 100 classes are composed of 20 super classes, and each super class possesses 5 child classes. The images in the CIFAR-100 dataset have a size of $32\times32$ like CIFAR-10. 
\item SVHN: SVHN (Street View House Numbers) dataset is composed of 630,420 RGB digital images with a size of $32\times32$, including a training set with 73,257 images and a test set with 26,032 images. 
\item MNIST: MNIST is a 10 class dataset of 0 $\sim$ 9 handwritten digits,including 60,000 training images and 10,000 test images in total. Each sample is a size $28\times28$ gray-scale image. 
\item KMNIST: Kuzushiji-MNIST (KMNIST) dataset is composed of 10 classes of cursive Japanese characters (namely, ``Kuzushiji'').  Each sample is a gray-scale image of $28\times28$ size. This dataset includes 60,000 training images and 10,000 test images in total. 
\item Fashion-MNIST: Fashion-MNIST (FMNIST for short) is a 10 class dataset of fashion items:T-shirt/top, Trouser, Pullover, Dress, Coat, Scandal, Shirt, Sneaker, Bag and Ankle boot. Each sample is a gray-scale image of $28\times28$ size. This dataset includes 60, 000 training images and 10,000 test images in total.
\end{itemize}

\subsection{Training Details}
The models utilized mainly in the experiments were designed based on the VGGNet-16. We added batch normalization layer after each convolutional layers to alleviate the issue of ``Internal Covariance Shift’’ and accelerate the convergence of the model. In addition, we removed the final three fully connected layers and utilized the global average pooling instead. Besides, the convolution layers with step-size were employed to conduct down-sampling operation instead of utilizing pooling method~\cite{cai2021study}. All the learned parameters were initiated with Xavier initialization method. We utilized the cross-entropy loss as the loss function. The training algorithm employed mini-batch stochastic gradient descent with a momentum term of 0.95 and weight decay coefficient of 0.0005, the batch size was set to be 64. All experimental models have been trained with NVIDIA GeForce GTX 2080Ti GPU.

For datasets CIFAR-10 and CIFAR-100 , the models were trained for 100 epochs. For others, it was 40 epochs. We utilized the multi-step learning rate strategy during training. For all datasets, the initial learning rate was set to 1e-3, and then was changed to 1e-4 during the last 20 epochs. For the training data of CIFAR-10 and CIFAR-100 data sets, we adopt the method of data augmentation that that first padding 4 circles of zero pixels around the image, randomly cropping the padded image to its original size, and then flipping the image horizontally by the probability of 50\%.

\subsection{Compared approaches}
We carry out 11 contrast experiments for every benchmark dataset, and the experimental results are shown in the later section. These eleven methods are listed in Table.~\ref{t2}. The mark “Normal” represents a normal method without the Interflow algorithm; S0\textasciitilde9 stand for 10 different approach to implementing the Interflow algorithm. 

To explain more clearly, methods S0\textasciitilde4 each contains 4 branches. Method S0 is employed with fixed weights of 0.1, 0.2, 0.3 and 0.4 to the corresponding branch manually. For method S1,  different categories of each branch share the same weight and it enables the model to learn 4 weights automatically with random initialization. For method S2, the weight of each branch is manually initialized to 0.1, 0.2, 0.3 and 0.4, the model is made to learn and update 4 weights independently. S3 is a method that has a total of 4 branches, different categories in each branch are assigned with different weights and makes the model learn a total of 40 weights independently; S4 bears resemblance to method S3 but initializes weights of 0.1, 0.2, 0.3 and 0.4 to each branch and makes the model learn and update 40 weights independently. 

For method S5\textasciitilde 9, they are similar to method S0\textasciitilde 4 accordingly but for that they possess 7 branches. In addition, when the mark ``Initialization’’ is True, the manually initialization weights are set as 0.1 for the first four branches and 0.2 for the last three branches. 


It is worth mentioning that the fully connected layer is shared by the outputs of different stages when the numbers of input channels are the same. As a matter of fact, experimental results indicated that training fully connected layers for different branches respectively would decrease the final prediction accuracy. 
\begin{table}[ht]
   \caption{Details of various methods.}
    \label{t2}
   \centering
   \scalebox{0.95}{
    \begin{tabular}{|c|c|c|c|c|c|}
    \hline
    \textbf{Method} & \textbf{Interflow} & \textbf{Branches} & \textbf{Shared} & \textbf{Initialization} & \textbf{Learned}  \\
    \hline
    Normal & False  & - & - & - & - \\
    \hline
    S0 & True  &4 &True &True &False \\
    \hline
    S1 & True &4 &True &False &True \\
    \hline
    S2 & True  &4 &True &True &True \\
    \hline
    S3 & True &4 &False &False &True \\
    \hline
    S4 &True  & 4&False &True &True \\
    \hline
    S5 & True&7&True &True &False \\
    \hline
    S6 & True &7 &True &False &True \\
    \hline
    S7 & True &7 &True &True &True \\
    \hline
    S8 & True &7 &False &False &True \\
    \hline
    S9 & True&7 &False &True&True \\
    \hline
    \end{tabular}
    }
 \end{table}

\begin{table*}[ht]
   \caption{Test accuracy of various methods on different benchmark datasets.}
    \label{t3}
   \centering
    \begin{tabular}{|c|c|c|c|c|c|c|}
    \hline
    \textbf{Method} & \textbf{CIFAR-10} & \textbf{CIFAR-100} & \textbf{SVHN} & \textbf{MNIST} & \textbf{KMNIST} & \textbf{F-MNIST} \\
    \hline
    Normal & 91.12 &67.78 &93.91 &99.39 &96.61 &93.18 \\
    \hline
    S0 & 91.75 &68.95 &94.23 &99.50 &97.12 &93.38 \\
    \hline
    S1 & 92.04 &68.68 &94.17 &99.54 &97.08 &93.55 \\
    \hline
    S2 & 91.74 &68.45 &94.32 &\textbf{99.58} &97.14 &93.32 \\
    \hline
    S3 & \textbf{92.12}  &68.53 &94.25 &99.57 &97.10 &93.36 \\
    \hline
    S4 & 91.93 &\textbf{69.05} &94.30 &99.55 &\textbf{97.17} &\textbf{93.64} \\
    \hline
    S5 & 91.75 &68.11 &94.20 &99.54 &97.11 &93.41 \\
    \hline
    S6 & 91.97 &68.31 &\textbf{94.34} &99.56 &97.10 &93.43 \\
    \hline
    S7 & 91.68 &68.23 &94.22 &99.57 &97.10 &93.38 \\
    \hline
    S8 & 91.64 &68.46 &94.24 &99.54 &97.12 &93.57 \\
    \hline
    S9 & 91.81 &68.44 &94.27 &99.56 &97.16 &93.47 \\
    \hline
    \end{tabular}
 \end{table*}
 
\subsection{Experimental results and analysis}
The test accuracy of each method on various benchmark datasets is shown in Table.~\ref{t3}. It is observed that the models utilizing the Interflow algorithm  outperform the original models on each benchmark dataset significantly, especially on the datasets CIFAR-10 and CIFAR-100 which are more difficult for models to recognize. The experimental results validate that the Interflow algorithm can improve the representation and generalization ability of the original model significantly.

According to Table.~\ref{t3}, soft attention always performs better than hard attention, thanks to the learned parameters. However, more learned parameters did not mean better performance when comparing method S2 and S4 on the easier datasets, SVHN, MNIST and FMNIST. What's more, providing initial weights for these learned parameters can obtain slightly better results. For the case of branch numbers, more interflow branches did not guarantee higher test accuracy, comparing method S6 with S1, S7 with S2, S8 with S3, and S9 with S4. Overall, the difference between different methods is not significant. 



\begin{figure*}[htbp]
\centering
\subfigure[Test accuracy]{
\includegraphics[height=0.25\textwidth]{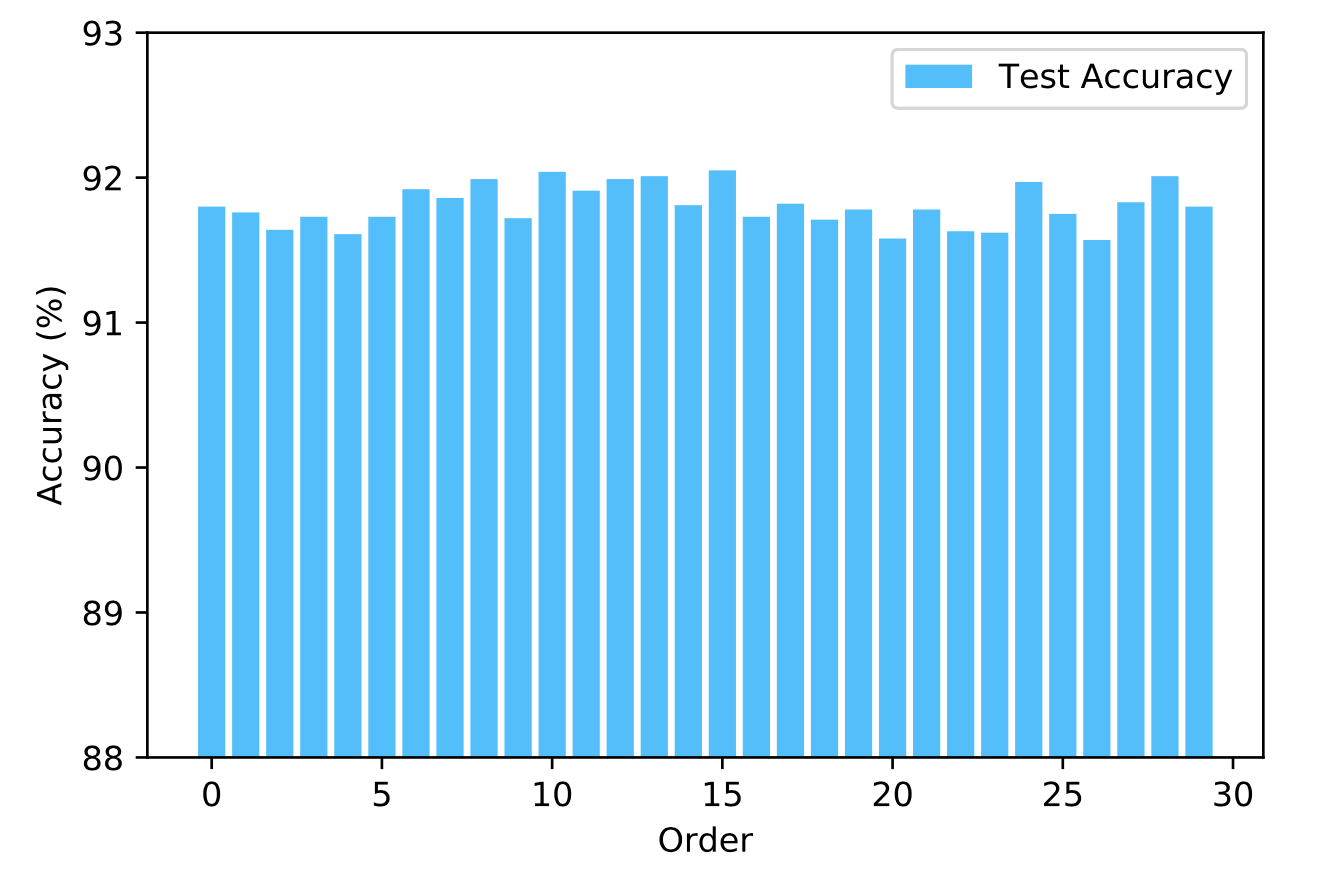}
}
\subfigure[Learned weights of 4 branches]{
\includegraphics[height=0.25\textwidth]{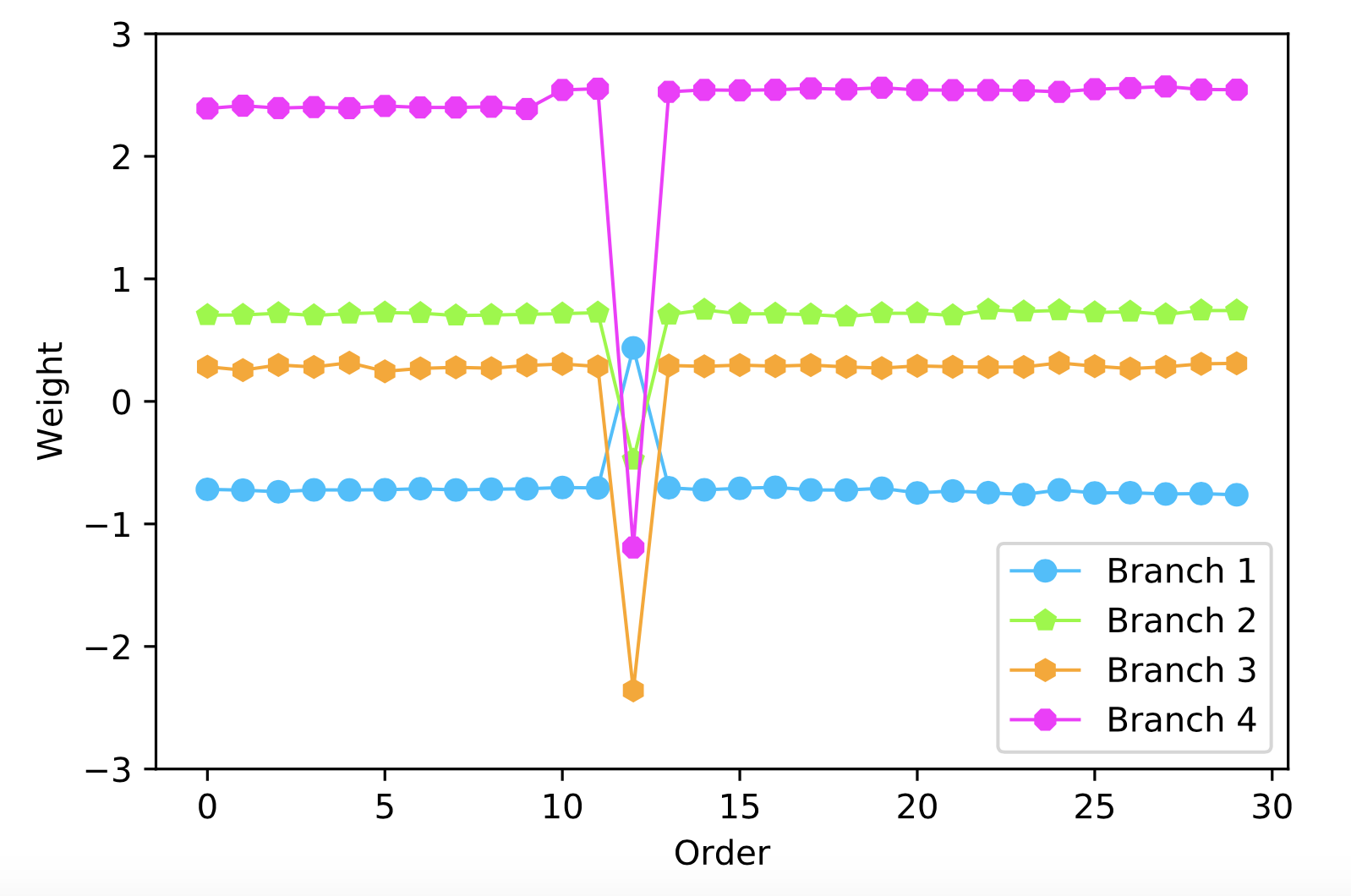}
}
\caption{Test accuracy (a) and the final learned weights (b) of each branch obtained by testing S1 method on CIFAR-10 dataset for 30 times. The abscissa is the order of the 30 experiments.}
\label{f4}
\end{figure*}

\begin{table}[ht]
   \caption{Average values and variances of test accuracy and learned weights including or excluding the 13-th experiment.}
    \label{t4}
   \centering
   \scalebox{0.90}{
    \begin{tabular}{|l|c|c|c|c|c|}
    \hline
    \textbf{Metrics} & \textbf{Accuracy (\%)} & \textbf{Branch1} & \textbf{Branch2} & \textbf{Branch3} & \textbf{Branch4}  \\
    \hline
    Mean &91.81&	-0.6881	&0.6791&	0.1978&	2.3720  \\
    \hline
    Var & 0.020	& 0.0439&	0.0462&	0.2256&	0.4428\\
    \hline
    Mean (-13) &91.80&	-0.7269&	0.6959&	0.2860	&2.4949 \\
    \hline
    Var \ \ \ (-13) & 0.019&	0.0003&	0.0143&	0.0003&	0.0049 \\
    \hline
    \end{tabular}
    }
 \end{table}
 
\begin{table*}[ht]
   \caption{Average values of these 40 weights learned in method S3 on CIFAR-10 dataset.}
    \label{t6}
   \centering
    \begin{tabular}{|c|c|c|c|c|c|c|c|c|c|c|}
    \hline
    \textbf{Branch} & \textbf{Cls 1}&\textbf{Cls 2} & \textbf{Cls 3}& \textbf{Cls 4}& \textbf{Cls 5}& \textbf{Cls 6}& \textbf{Cls 7}&\textbf{Cls 8} & \textbf{Cls 9}&\textbf{Cls 10} \\
    \hline
    Branch 1 &-0.3885&	0.5546	&0.3028&	-0.2134&	-0.2081&	-0.0911&	-0.0247&	-0.2253&	-0.0394&	-0.5765 \\
    \hline
    Branch 2 &0.6107&	0.0505&	0.3650&	0.6947&	0.0509&	-0.9258&	0.7384&	0.7112	&-0.7594&	-0.1169 \\
    \hline
    Branch 3 &0.4908&	-1.0578&	0.8717&	0.2350&	0.7161&	0.8392&	0.7092&	0.5764&	0.7152&	-1.2599 \\
    \hline
    Branch 4 &1.1658&	1.0862&	-1.0918& 1.0128& -1.1372& 0.8636& -1.1506& 1.1135& -1.0614&	0.7656 \\
    \hline
    \end{tabular}
 \end{table*}

\begin{table*}[!t]
\renewcommand{\arraystretch}{1.3}
\caption{Test accuracy of extremely deep CNN models with/without Interflow, and the weights learned.}
\label{t7}
\centering
\scalebox{0.85}{
\begin{tabular}{|c|c|c|c|l|}
\hline
\textbf{Depth} & \textbf{Normal} & \textbf{Interflow} &\textbf{Branches} &\textbf{Learned weights}\\
\hline
40&	91.03&	91.60& 5&	-2.4543, -1.1796, 0.7401, 0.6060, 0.3428 \\
\hline
50&	82.21&	91.55&7 &	2.7535, 0.5910,-0.4226, -0.2770, -0.2512, -0.1325, 0.0196 \\
\hline
60&	31.48&	90.39&9&	2.9065, -0.2477, -0.4238, 0.0191, -0.1099, -0.3008, -0.3012, -0.1194, -0.3278 \\
\hline
70&	29.88&	90.36&11&	2.8246, -0.5821, -0.0617, -0.1784, -0.3335, -0.3279, -0.0849, -0.0069, 0.0001, -0.0058, -0.0276\\
\hline
80&	35.29&	88.05&13&	-2.6898, 0.5479, 0.2242, 0.2640, 0.1101, 0.2565, -0.0210, 0.0209, 0.1539, -0.0346, 0.0012, -0.0534 0.7578 \\
\hline
90& 	10.00&	87.55&15&	3.8587, -0.4778, -0.2281, -0.0808, 0.0410, -0.0608, -0.0877, 0.0568, -0.0199, 0.1128, -0.4170, -0.0108, -0.0486, -2.0337, 0.1874 \\
\hline
100 &	10.01& 85.91&15&	-2.0836, 0.3801, 0.0767, 0.1188, -0.0165, -0.0480, -0.0364, -0.0364, -0.0212, 0.0002, 0.0034, -0.0327, -0.1295, -0.0127, 1.3403 \\
\hline
\end{tabular}
}
\end{table*}

\begin{table}[!t]
\renewcommand{\arraystretch}{1.3}
\caption{Test accuracy obtained by running S3 method for ten times on CIFAR-10 dataset.}
\label{t5}
\centering
\begin{tabular}{|c|c|}
\hline
Order & Test Accuracy (\%)\\
\hline
0 & 91.75 \\
\hline
1& 91.89 \\
\hline
2& 91.77 \\
\hline
3& 91.87 \\
\hline
4& 91.92 \\
\hline
5& 91.95 \\
\hline
6& 91.95 \\
\hline
7& 91.92 \\
\hline
8& 91.75 \\
\hline
9& 92.12 \\
\hline
Mean & 91.89 \\
\hline
Var & 0.01 \\
\hline
\end{tabular}
\end{table}

\subsection{Algorithm Analysis}
We tested S1 method on CIFAR-10 dataset for 30 times. The test accuracy and the final learned weights of each branch are exhibited in Fig.~\ref{f4}. The first two lines in Table.~\ref{t4} are respectively the average values and variances of accuracy and learned weights of each branch in total 30 experimental tests. It is noticeable that variances are small. Furthermore, since only the result of the 13-th experiment greatly deviates from the mean value, it can be considered to represent another local minimum. If the result of the 13-th experiment is eliminated, the variances are close to zero, which means the weights in each Interflow branch converge stably and break free from fluctuations. The average values and variances of accuracy and learned weights excluding the 13-th experiment are listed in the last two lines of Table.~\ref{t4}.

Consequently, we analyze the weight parameters currently. The highest weights emerge in the 4-th branch which stands for the deepest stage. Regarding to the contribution to the prediction results, the deepest convolutional layer can extract more abstract and discriminative features and learn richer semantic information. As a result, it plays the decisive role in the final prediction; On the other hand, it also indicates that there is neither serious over-fitting nor network degeneration in the network at this time. In addition, with respect to the convergence of each layer of the model, the deep convolutional layers can converge faster, and the problem of improper selection of the hyper-parametric learning rate can be alleviated if larger weights to the deeper convolutional layers are added. However, in the shallow stage, there are weights of negative phase, which enables the shallower convolutional layers to continuously carry out learning after the initial convergence. From the perspective of back propagation, the parameter update of the shallower weight layer is not only propagated by the loss of the higher weight layer, but also directly propagated by the total loss, which reduces the gradient vanishing problem naturally.

In the 13-th experiment, the experimental result is comparatively excellent, but its weights in each branch deviate greatly from the mean value, which can be considered that this experiment has converged to another local minimum area. It is found that the weights of deeper branches have negative phases, which is actually caused by the negative phase of the parameters in the fully connected layers. At this time, it is supposed to analyze the absolute value of the weight. We find that the feature information in the third stage occupies the largest weight ratio. From the data in Table.~\ref{t1}, it can be observed that the test accuracy bears very slight difference between the model with 11 convolutional layers and the model with 13 convolutional layers. These two baseline models with different depths have almost the same fitting ability and generalization performance, indicating the strong feature extraction ability of relatively middle layers. As a result, it is reasonable for the third stage branch in 13-th experiment to learn a significant weight.

In addition, we tested S3 method which makes the model learn individual weights for different categories in each branch on the CIFAR-10 dataset for ten times. The test accuracy obtained are shown in Table.~\ref{t5}. Table.~\ref{t6} exhibits the average values of these 40 weights learned. Some classes in shallower branches are attributed relatively strong weights like Cls7 and Cls9 in branch2, which illustrates that shallower branches may contribute a lot for the final prediction in some classes. Moreover, the learned weights are consistent to the principle that the deeper the branch is, the larger the weights are. However, this principle only establishes when the depth of the CNN model is moderate. For the circumstance of the extremely deep CNNs, it will be not the case.  

\subsection{Case of Extremely Deep CNNs}
When the CNN reaches an extreme depth, the network degradation phenomenon will emerge. However, it is not an overfitting problem since the training loss and test loss of the model are both very high at this time. The critical cause is attributed to the significant decrease of the freedom degree for model effective fitting. In short, there are only a minor number of hidden units in each convolutional layer that are sensitive to different inputs, thereby changing their activation values accordingly. Unfortunately, most of hidden units have the same response to different stimulation. At this time, the neural network can hardly learn useful feature information. As a byproduct of the Interflow algorithm, we apply Interflow to an extremely deep CNN, which is supposed to endow zero or a minimum value to the branches of very deep stages through taking advantage of the attention mechanism. As a result, the effective layers of the network can be adaptively reduced  to an appropriate depth and the degradation problem of the network can be addressed. Table.~\ref{t7} shows the CNN models of different deep depths with or without the Interflow algorithm measured on the CIFAR-10 benchmark dataset. The weights of different stages learned by models with Interflow algorithm are also displayed in Table.~\ref{t7}. 
It is noticeable that when the network depth reaches 60 layers, the CNN model with the Interflow algorithm can still achieve a favorable accuracy in the test set (the problem of network overfitting is not significant, which can be ignored), while the ordinary CNN model drops significantly in test accuracy (the actual training accuracy is also very low). When observing the learned weights, we find that the weights are relatively large in branches of the shallower stage, while the weights are extremely small in branches of the shallower stage, indicating that the attention mechanism with the Interflow algorithm plays a positive role. When the network depth exceeds 80 layers, the accuracy of the ordinary CNN model in test set is about 10.00\%, indicating that it completely fails to learn any effective feature information. However, the CNN model with our Interflow algorithm can still learn more effective information, and its test accuracy is significantly higher than the ordinary CNN model. This phenomenon validates the strength of Interflow. With the observation of the learned branch weights, it is found that the shallowest stages learned the majority weights, which may contribute to the effectiveness of the attention module. Besides, the reason for the CNN with Interflow failing to obtain a preferable generalization ability may be that the branches at the deeper stages still learn too large weight for certain classes, thereby providing wrong contribution to the network's final prediction. The experimental results validate that, as a byproduct, Interflow can mitigate the network degradation problem for extremely deep CNN models.

\section{Conclusion}
This paper proposes the Interflow algorithm, which adds extra prediction branches and aggregates the feature information of these branches with attention mechanism to obtain the final prediction output. Interflow can lighten the difficulty of selecting network depth and force the intermediate layers to learn more distinguished features. It also possesses many other advantages as mentioned above. Experimental results indicate that the Interflow algorithm can lift the model representation ability as well as the accuracy of prediction.

\end{document}